\documentclass{article}

\usepackage[sglblindworkshop,final]{neurips_2026}
\workshoptitle{PhysUnderstand}

\usepackage[utf8]{inputenc}
\usepackage[T1]{fontenc}
\usepackage{amsmath,amssymb,amsfonts}
\usepackage{graphicx}
\usepackage{booktabs}
\usepackage{natbib}
\usepackage{hyperref}
\usepackage{url}
\usepackage{multirow}

\newcommand{\blfootnote}[1]{%
\begingroup
\renewcommand\thefootnote{}%
\footnotetext{#1}%
\endgroup
}

\usepackage[skip=4pt,font=small]{caption}
\title{How Much Physics Can Vision Learn from Audio?\\ Probing and Distilling Damping Structure}

\author{
  Zainab Afolabi \\
  Amazon Web Services \\
  \texttt{zafolabi@amazon.co.uk}
  \And
  Vadim Selyutin *\\
  Amazon Web Services \\
  \texttt{seliutin@amazon.ae}
  \And
  Zhao Gao * \\
  Amazon Web Services \\
  \texttt{orangez@amazon.co.uk}
  \And
  Babs Khalidson *\\
  Amazon Web Services \\
  \texttt{khalbabs@amazon.co.uk}
  \And
  Mohamed Ghoneim \\
  Amazon Web Services \\
  \texttt{ghoneimm@amazon.co.uk}
}

\begin{document}

\maketitle
\blfootnote{$^{*}$Equal contribution.}

\begin{abstract}
Physical understanding is crucial for decision making, yet vision models struggle to infer properties such as damping that govern how materials behave. Audio encodes such properties directly, because an object's acoustic response to impact reveals its damping through measurable decay. We probe frozen encoders for the quality factor $Q$, a dimensionless measure of how many oscillations an object completes before its ring dies away, and report how well each embedding orders impacts by $Q$ using the Spearman rank correlation $\rho$, for which $\rho=1$ is a perfect ordering and $\rho=0$ an unrelated one. Audio embeddings reach $\rho=0.73$ and retain most of that advantage on materials withheld from training, while a single video frame taken at the moment of impact, encoded by CLIP, supports only $\rho=0.28$. Having established where the signal lives, we ask how much of that gap vision can recover. We distill a frozen audio teacher into a frozen CLIP backbone through a trained projection head and find that this recovers a useful fraction of the audio advantage. Because pooled correlations on this task are dominated in practice by category identity, we also evaluate under a leave-one-material-out protocol in which the held-out material is withheld from the projection head as well as the probe. Distillation with a four-frame input improves the correlation there from 0.234 to 0.338 (+0.053 over the matched four-frame baseline, 14 of 17 materials), a gain that no category-level strategy can produce. Widening the visual input to four frames spanning the impact supplies half of that improvement at no acoustic cost, which indicates that part of what a single frame misses is visible dynamics.
\end{abstract}

\section{Introduction}

A robot clearing a table observes two cups that look alike, one ceramic and one plastic. A vision model might correctly label their materials, but this categorical distinction does not reveal how they behave, since the ceramic cup shatters when dropped while the plastic cup bounces. Without physical grounding, vision cannot predict these divergent futures.

This appearance--reality gap pervades modern computer vision. Large vision models achieve strong recognition performance, but what they encode is how things look rather than how things behave. Physics benchmarks confirm this gap. PhysBench~\citep{chow2025physbench} found that GPT-4o achieves only 46.99\% on physical dynamics (49.49\% overall, against 95.87\% for humans), and Physion++~\citep{tung2023physionpp} showed that all models remain well below human performance on latent properties such as mass, friction, or elasticity.

Audio captures what appearance does not. An object's acoustic response to impact is governed by physical laws: stiffness, density, and geometry set the resonant frequency, and internal friction, radiation, and mounting set the damping. We confirm the gap directly. Frozen audio embeddings predict the quality factor $Q$ at $\rho = 0.73$, while a single CLIP-encoded video frame taken at the moment of impact supports only $\rho = 0.28$. Here $Q$ is a dimensionless measure of how many oscillations a struck object completes before its ring dies away, so a highly resonant glass has a large $Q$ and a heavily damped cloth a small one. We write $\rho$ for the Spearman rank correlation between predicted and true values: $\rho=1$ means the model orders impacts by $Q$ perfectly, and $\rho=0$ means its ordering carries no information about the truth. We use $\rho$ throughout because we care about whether a model can rank objects by how much they ring, not about the absolute scale of its predictions. Our question is the narrower one of how far a frozen visual encoder can be pushed toward the audio number.

Our answer is that it can be pushed a useful fraction of the way. We distill from a frozen audio teacher into a frozen CLIP encoder, training a projection head as the only new component, and measure the outcome under a protocol in which knowing the material category earns nothing: correlations are computed within a material that was held out from the projection head and from the probe alike. Distillation improves prediction there for most of the 17 materials. Part of this improvement needs no audio at all, since giving the encoder four frames spanning the impact, so that it can see the object deform and recoil, supplies about half. Most of the audio signal nonetheless stays out of reach, which we take to mark the interesting part of the problem rather than a settled question.

\subsection{Contributions}
\begin{itemize}
  \item We propose probing frozen encoders for a continuous physical property, the quality factor $Q$, rather than for the categorical material labels used in prior evaluations. Our measurements show that audio embeddings predict $Q$ at $\rho=0.73$ whereas a single CLIP frame reaches 0.28, that this audio advantage persists for materials withheld from training ($\rho \approx 0.60$), and that every visual encoder we test is near chance at ranking objects within wood, ceramic, and plastic and reaches at most 0.36 within metal.
  \item We show that part of the visual shortfall is temporal rather than material. Four frames spanning the impact raise vision-only prediction on unseen materials by 0.051 without any acoustic supervision, which suggests that some of what a single frame misses is visible dynamics rather than hidden material structure.
  \item We propose distilling a frozen audio teacher into a frozen visual encoder through a trained projection head, and our experiments demonstrate that this moves a useful fraction of the signal across. The distilled CLIP encoder reaches $\rho=0.44$ on $Q$ and, with a four-frame input, improves prediction from 0.234 to 0.338 on materials withheld from the head as well as the probe (+0.104, 15 of 17 materials, $p=0.002$; +0.053 over the matched four-frame baseline, 14 of 17 materials).
  \item We report controls that establish where this gain comes from. A capacity ablation, a shuffled-teacher control, and a material-label reference together attribute most of the improvement to acoustic correspondence rather than to added parameters or category recognition. We also document the experimental choices that inflate such measurements when these controls are absent.
\end{itemize}

\section{Related Work}

Impact sounds encode material properties that appearance cannot reveal. \citet{owens2016visually} established this connection with the Greatest Hits dataset, demonstrating that visual models can learn to predict impact sounds. RealImpact~\citep{clarke2023realimpact} extends this with force profiles and calibrated measurements. ObjectFolder~\citep{gao2023objectfolder} quantifies the relative contributions of vision, audio, and touch for material understanding, finding that audio often provides complementary information. In robotics, audio frequently outperforms vision for physical property estimation: \citet{thankaraj2023sounds} show that auditory feedback improves manipulation, and SonicSense~\citep{liu2024sonicsense} demonstrates object perception through acoustic vibration. We connect this evidence to representation learning: if audio encodes physics, can this structure be distilled into vision?

Joint audio-visual embeddings have evolved from contrastive methods~\citep{arandjelovic2017look, owens2018audio} to masked autoencoders~\citep{gong2023cavmae, huang2023mavil}. PE-AV~\citep{vyas2025pushing} scales to 100M pairs, achieving state-of-the-art results on audio-visual retrieval. ImageBind~\citep{girdhar2023imagebind} aligns six modalities through image-paired data. Prior work evaluates these models on retrieval and classification, but, to our knowledge, none probes what frozen audio-visual embeddings capture about \emph{physical} properties, or whether this knowledge can transfer to vision alone.

Recent benchmarks reveal systematic failures in physical reasoning. Beyond PhysBench~\citep{chow2025physbench}, Physion~\citep{bear2021physion} shows that models fall short of humans at predicting physical outcomes, Physion++~\citep{tung2023physionpp} extends this to outcomes that depend on latent properties, and CLEVRER~\citep{yi2019clevrer} and IntPhys~\citep{riochet2021intphys} probe causal and intuitive physics understanding. These benchmarks focus on dynamics prediction, and we complement them by probing continuous material properties.

Audio and vision have long supervised one another: ambient sound can train visual representations~\citep{owens2016ambient}, and visual networks can train sound representations~\citep{aytar2016soundnet}. Knowledge distillation transfers representations across architectures~\citep{hinton2015distilling} or modalities: \citet{gupta2016cross} transferred supervision from labeled RGB images to depth and optical-flow representations. Our work differs in targeting \emph{specific physically relevant structure} rather than general representations, and takes a step toward making acoustically grounded properties partially accessible to vision alone.

\section{The Quality Factor: A Continuous Physical Property}

\subsection{Why \texorpdfstring{$Q$}{Q}?}

The quality factor $Q = \pi f_0 / \beta$ combines resonant frequency ($f_0$) and damping rate ($\beta$) into one dimensionless scalar. The damping rate $\beta$ sets how quickly a ringing object falls silent in absolute time, whereas $Q$ counts how many oscillation cycles it completes before decaying, so a high-pitched glass can fall silent sooner than a lower-pitched metal bowl while having the higher $Q$.

We target this compound quantity for three reasons. Being dimensionless, $Q$ is comparable across objects regardless of size or pitch, so a wine glass and a steel beam can share a $Q$ despite frequencies orders of magnitude apart. It also varies continuously within material categories: a thin metal sheet and a solid metal block are both ``metal'' yet differ substantially in $Q$, so predicting it demands more than category recognition. Finally, it has direct physical consequences, determining whether an object rings, bounces, or absorbs energy on impact. These properties make $Q$ a stricter test of physical grounding than frequency, which is largely predictable from visible geometry, or the damping rate alone, which is confounded with object size.

\subsection{Extraction Pipeline}
\label{sec:extraction}

We extract $Q$ from audio recordings in the Greatest Hits dataset~\citep{owens2016visually}, which contains 977 videos of objects being hit with a drumstick. Our copy carries annotations for 975, totaling 28,444 individual impacts across 17 material categories.

\begin{figure}[t]
  \centering
  \includegraphics[width=\textwidth]{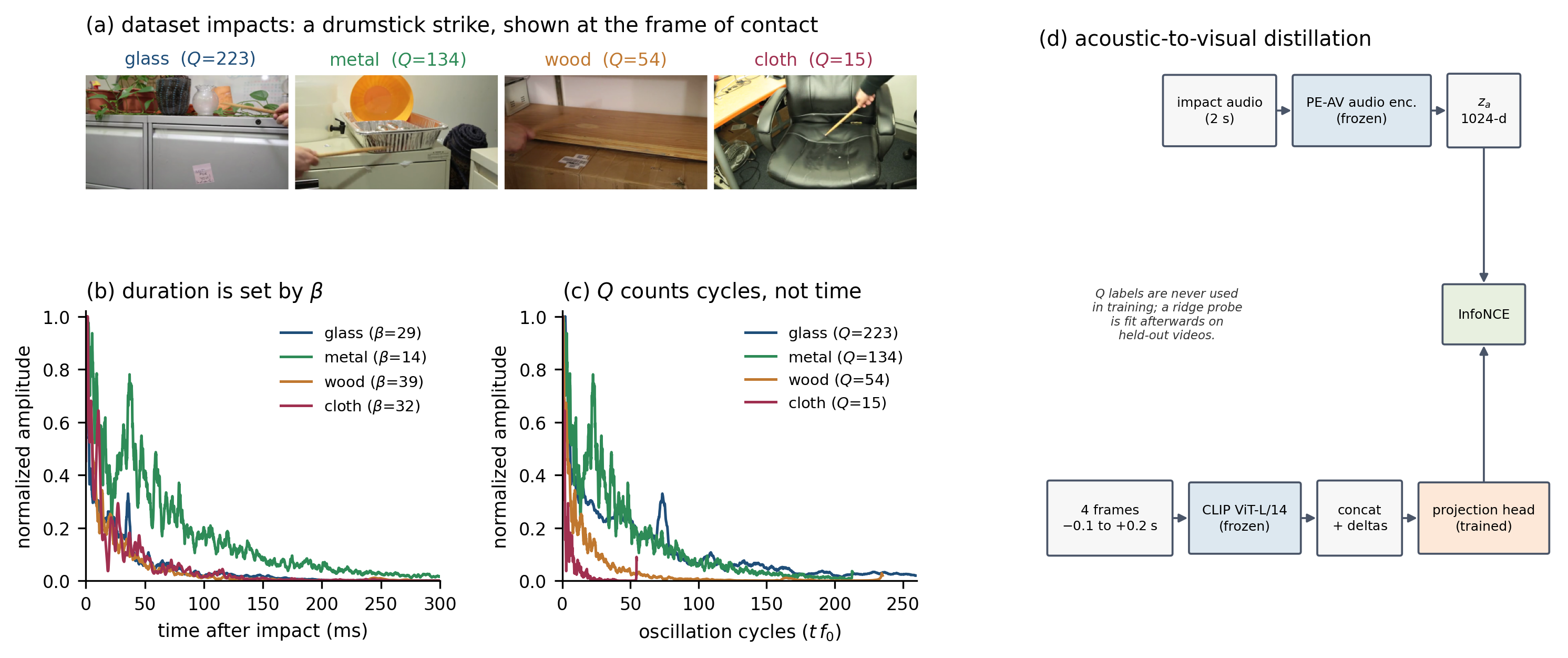}
  \caption{\textbf{What $Q$ measures, and how we transfer it.} (a) Example impacts from the Greatest Hits dataset: a drumstick strike shown at the frame of contact, with example $Q$ values ranging from 223 (glass) to 15 (cloth); per-material medians are given in Section~\ref{sec:validation}. (b) Hilbert envelopes plotted against time show that absolute duration is governed by the damping rate $\beta$: the metal bowl ($\beta = 14$~s$^{-1}$) rings longest even though glass has the higher $Q$. (c) Plotted against elapsed oscillation cycles $t f_0$, the ordering follows $Q$, with cloth and wood falling silent within tens of cycles while glass and metal persist past two hundred. $Q$ therefore measures the quality of resonance rather than its duration. (d) The distillation setup: both encoders stay frozen and only the projection head is trained, with a symmetric InfoNCE objective. No $Q$ label enters training, and the ridge probe used for evaluation is fit afterward on held-out videos.}
  \label{fig:method}
\end{figure}

Figure~\ref{fig:method} illustrates both the quantity and the method. For each annotated impact, we take the magnitude of the analytic signal (the Hilbert envelope) to obtain an amplitude envelope $A(t)$ and fit the decay model $A(t) = A_0 e^{-\beta t} + C$ by nonlinear least squares, with $\beta$ bounded to $[0.1, 1000]$~s$^{-1}$ and the additive constant $C$ absorbing background noise. The resonant frequency $f_0$ is the dominant peak of the power spectrum above 50~Hz, a floor that avoids low-frequency room noise. Retaining only fits with $R^2 \geq 0.8$ leaves 17,958 usable impacts, 63\% of the total. Rejections concentrate on granular and scattering surfaces, where the response has no single resonant mode to fit: water retains 24\% of its annotated impacts, gravel 33\%, and dirt 36\%, against 83\% for cloth and 91\% for glass. We return to this selection effect in Section~\ref{sec:limitations}.

\subsection{Validation}
\label{sec:validation}

The extracted values match physical expectation. Median $Q$ rises monotonically from the soft, absorptive materials to the resonant ones: cloth 15, carpet 23, wood 55, plastic 89, metal 134, ceramic 327, glass 412. This ordering is consistent with the extraction capturing physically meaningful variation rather than recording artifacts.

\section{Physical Structure in Audio-Visual Embeddings}

\subsection{Models Evaluated}

We probe four frozen encoders. Two are audio-visual: PE-AV~\citep{vyas2025pushing}, trained contrastively on 100M audio-visual pairs and exposing separate audio, video, and joint embeddings, and CAV-MAE~\citep{gong2023cavmae}, a masked autoencoder trained on AudioSet~\citep{gemmeke2017audioset} with both reconstruction and contrastive objectives. Two are vision-only: CLIP ViT-L/14~\citep{radford2021clip}, trained on 400M image--text pairs, and DINOv2 ViT-L/14~\citep{oquab2023dinov2}, trained by self-distillation on curated images without language supervision. The latter pair represents the two dominant recipes for general-purpose visual features, and neither has any acoustic exposure. Video inputs are single frames taken at the annotated moment of impact, each encoded as an image, and audio inputs are 2-second clips centered on the impact.

\subsection{Probing Protocol}
\label{sec:protocol}

We train ridge regression ($\lambda = 1.0$) from frozen embeddings to $\log(Q + 1)$ and evaluate with Spearman $\rho$. Pooled correlations carry 95\% bootstrap intervals over samples (1,000 resamples). Leave-one-material-out figures are means over the 17 held-out materials, and their intervals resample the materials rather than the samples (10,000 resamples), which is why they are wider.

\subsection{Results}

\begin{table}[t]
\centering
\small
\caption{\textbf{Audio embeddings capture the quality factor $\boldsymbol{Q}$, while single-frame vision largely does not.} Spearman $\rho$ between ridge-probe predictions from frozen embeddings and ground-truth $Q$, on the held-out video split.}
\label{tab:qfactor}
\begin{tabular}{@{}llrl@{}}
\toprule
\textbf{Model} & \textbf{Modality} & $\boldsymbol{\rho}$ & \textbf{95\% CI} \\
\midrule
PE-AV & Audio & \textbf{0.727} & [0.707, 0.746] \\
CAV-MAE & Audio & 0.701 & [0.681, 0.721] \\
PE-AV & Joint & 0.696 & [0.676, 0.717] \\
CAV-MAE & Joint & 0.608 & [0.583, 0.633] \\
\midrule
PE-AV & Video & 0.362 & [0.333, 0.393] \\
CLIP & Video & 0.281 & [0.247, 0.313] \\
DINOv2 & Video & 0.271 & [0.242, 0.303] \\
CAV-MAE & Video & 0.151 & [0.121, 0.181] \\
\bottomrule
\end{tabular}
\end{table}

Table~\ref{tab:qfactor} reports the result. The gap between audio ($\rho = 0.73$) and CLIP vision ($\rho = 0.28$) quantifies physical structure that frozen single-frame visual encoders do not linearly expose. Co-training on audio does not by itself place that structure in the visual pathway: PE-AV's video branch reaches 0.362 and CAV-MAE's only 0.151, both far below their own audio branches, and the two models differ in architecture, data scale, and objective, none of which this comparison controls. We therefore treat the audio results as motivation for pushing structure into a frozen visual encoder deliberately, rather than as evidence that co-training supplies it for free.

\section{Generalization Across and Within Materials}

\subsection{Leave-One-Material-Out (LOMO)}

High correlation on the full dataset could reflect material-specific memorization. We test generalization: train on 16 materials and evaluate on the held-out 17th.

\begin{table}[t]
\centering
\small
\caption{\textbf{Physical structure generalizes to unseen materials.} Mean over the 17 held-out materials, with intervals bootstrapped over materials.}
\label{tab:lomo}
\begin{tabular}{@{}llrl@{}}
\toprule
\textbf{Model} & \textbf{Modality} & \textbf{LOMO $\boldsymbol{\rho}$} & \textbf{95\% CI} \\
\midrule
PE-AV & Audio & \textbf{0.600} & [0.528, 0.658] \\
CAV-MAE & Audio & 0.595 & [0.538, 0.645] \\
PE-AV & Joint & 0.572 & [0.505, 0.630] \\
CAV-MAE & Joint & 0.492 & [0.441, 0.541] \\
\midrule
DINOv2 & Video & 0.239 & [0.177, 0.299] \\
CLIP & Video & 0.227 & [0.167, 0.289] \\
CAV-MAE & Video & 0.206 & [0.148, 0.264] \\
PE-AV & Video & 0.197 & [0.144, 0.245] \\
\bottomrule
\end{tabular}
\end{table}

Audio embeddings predict $Q$ for \emph{unseen} materials ($\rho \approx 0.60$; Table~\ref{tab:lomo}), and joint embeddings reach 0.49--0.57. This is physical generalization: a probe never trained on the held-out material still ranks its impacts by how much they ring, using only their audio embeddings.

\subsection{Within Materials, and Which Component of \texorpdfstring{$Q$}{Q}}
\label{sec:within}

Two checks ask whether these correlations are really about physics. The first computes $\rho$ within a single material category, where category identity supplies nothing. The modalities separate sharply: PE-AV audio reaches 0.67 on metal and 0.57 on wood, whereas every visual encoder, including PE-AV's video branch, is at or below 0.36 on metal and close to zero ($|\rho| \leq 0.19$) on wood, ceramic, and plastic. CLIP scores 0.06, 0.03, $-$0.14, and $-$0.19 on metal, wood, ceramic, and plastic. Baseline vision barely distinguishes physical variation among objects of the same material. Table~\ref{tab:within} in Appendix~\ref{app:within} gives the breakdown.

The second asks which component of $Q$ is responsible, since $Q = \pi f_0 / \beta$ compounds a spectral term with a damping term, and across our impacts $Q$ is strongly rank-correlated with $f_0$ ($\rho = 0.87$) and moderately anti-correlated with $\beta$ ($\rho = -0.54$). An encoder that merely recovered pitch would therefore score well. Probing the targets separately (Appendix~\ref{app:decompose}) shows that the damping term is the one audio captures best: PE-AV audio reaches $\rho = 0.729$ on $\beta$ against 0.700 on $f_0$, and the gap widens where category identity stops helping, with $\beta$ holding at 0.686 on unseen materials while $f_0$ falls to 0.485. Vision shows the opposite pattern at much lower levels, favoring $f_0$ when pooled (0.347 against 0.221), which is consistent with pitch being partly inferable from visible size. The reading we place on $Q$ therefore survives being tested directly, and $\beta$ is the sharper target.

\section{Acoustic-to-Visual Distillation}

\subsection{Setup}

The teacher is the frozen PE-AV audio encoder, which predicts $Q$ at $\rho = 0.73$. The student is CLIP ViT-L/14, also frozen, with a trainable three-layer MLP head (GELU, LayerNorm, 1024-dimensional output) as the only component that learns. Its input is four frames taken at $-0.1$, 0, $+0.1$, and $+0.2$~s around the annotated impact, each encoded independently and concatenated with their successive differences. Training minimizes a symmetric InfoNCE loss~\citep{oord2018cpc} between the projected visual embedding and the teacher's audio embedding ($\tau = 0.07$), with batch size 256 and AdamW~\citep{loshchilov2019adamw} at a learning rate of $10^{-4}$.

Because the space of plausible configurations is large, we fixed the feature variant and the number of epochs on a validation split drawn from 20\% of the training videos (which selected the four frames + deltas input, trained for two epochs) and used the test split for no selection, reporting three seeds. One detail of the data handling carries more weight than it may appear to. We use the dataset's video-level split throughout and train the head on training-split impacts only (22,793 of 28,444 impacts, from 780 of 975 videos), because the teacher's own embeddings predict $Q$ at $\rho = 0.73$: any impact seen during distillation has effectively had its target leaked, even though no $Q$ label is ever used.

\begin{figure}[t]
  \centering
  \includegraphics[width=\textwidth]{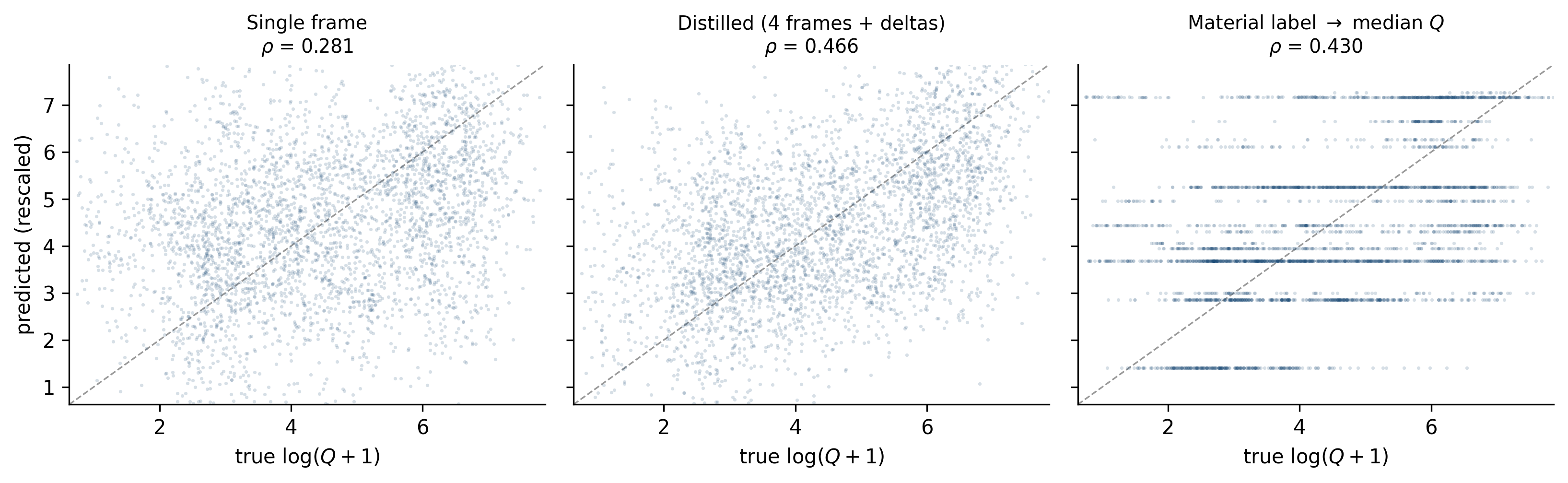}
  \caption{\textbf{Why pooled correlation is a weak measure of physical understanding.} Predicted against true $Q$ on the held-out split ($n = 3{,}510$), with predictions standardized to the target scale and the diagonal shown for reference. \emph{Left:} a single frame at impact supports only a weak trend. \emph{Middle:} after distillation, the cloud tightens appreciably. This panel shows the first, and highest, of the three seeds ($\rho = 0.466$, against the three-seed mean of 0.440 in Table~\ref{tab:distill}). \emph{Right:} mapping the true material label to its per-category median $Q$ produces just 17 distinct predictions, visible as horizontal bands, and still reaches $\rho = 0.430$. A model that cannot separate two metal objects at all therefore scores almost as well on this measure as our distilled encoder, which is why the argument rests on Section~\ref{sec:truelomo} instead.}
  \label{fig:scatter}
\end{figure}

\subsection{How Much of the Gap Is Temporal?}

Before introducing audio at all, we ask whether a single frame is the right visual input. A struck object deforms and recoils, and how quickly that motion subsides is what damping describes.

\begin{table}[t]
\centering
\small
\caption{\textbf{$\boldsymbol{Q}$ prediction from vision on the held-out video split, with and without distillation from the audio teacher.}}
\label{tab:distill}
\begin{tabular}{@{}lrll@{}}
\toprule
\textbf{Visual input / model} & $\boldsymbol{\rho}$ & \textbf{Interval} & \textbf{Type} \\
\midrule
\multicolumn{4}{@{}l}{\emph{No audio supervision}} \\
Single frame at impact & 0.281 & [0.247, 0.313] & 95\% CI \\
Four frames, concatenated & 0.299 & [0.269, 0.332] & 95\% CI \\
Four frames + deltas & 0.298 & [0.267, 0.329] & 95\% CI \\
\midrule
\multicolumn{4}{@{}l}{\emph{Distilled from the audio teacher}} \\
Four frames + deltas & 0.440 & [0.426, 0.466] & three-seed range \\
Material-label reference (Section~\ref{sec:controls}) & 0.430 & [0.400, 0.457] & 95\% CI \\
Teacher (PE-AV audio) & 0.727 & [0.707, 0.746] & 95\% CI \\
\bottomrule
\end{tabular}
\end{table}

Widening the window helps, but only modestly on the pooled measure: $\rho$ rises from 0.281 to 0.298 (0.299 without deltas), with overlapping intervals. The effect is clearer under the LOMO protocol of Section~\ref{sec:truelomo}, where the same change is worth +0.051 (0.234 to 0.285). We therefore treat the temporal contribution as modest, while noting that it costs no acoustic supervision whatsoever.

\subsection{Distillation Results}

Distillation raises vision-only $Q$ prediction to 0.440, from 0.298 with the same four-frame input and no teacher. This is marginally above the 0.430 obtainable from perfect material-category knowledge and still far below the audio teacher. Because the pooled measure is dominated in practice by category information, exceeding that reference by 0.01, which lies inside the three-seed range, is not conclusive (Figure~\ref{fig:scatter}), so we turn to a measure under which category identity earns nothing.

\subsection{Leave-One-Material-Out, with the Head Held Out}
\label{sec:truelomo}

LOMO is the discriminating test, because correlations are computed within a held-out material: a model that outputs a category's median $Q$ predicts a constant there and is scored zero, since Spearman $\rho$ is undefined for constant predictions. Two conditions must both hold for the test to mean anything. The material must be excluded from the probe, and it must also be excluded from the distillation head---otherwise the head has already learned that material's appearance-to-audio mapping and the probe merely reads it back. We refer to this stricter protocol as \emph{true LOMO}.

\begin{table}[t]
\centering
\small
\caption{\textbf{True LOMO: the held-out material is excluded from the projection head as well as the probe.} Mean over 17 materials and three seeds, with intervals bootstrapped over materials.}
\label{tab:truelomo}
\begin{tabular}{@{}lrll@{}}
\toprule
\textbf{Model} & \textbf{LOMO $\boldsymbol{\rho}$} & \textbf{95\% CI} & \textbf{vs.\ single frame} \\
\midrule
Baseline, single frame$^\dagger$ & 0.234 & [0.175, 0.289] & --- \\
Baseline, four frames + deltas$^\dagger$ & 0.285 & [0.200, 0.372] & +0.051 \\
Distilled, four frames + deltas & 0.338 & [0.249, 0.430] & +0.104 \\
\bottomrule
\end{tabular}
\par\vspace{2pt}
{\footnotesize $^\dagger$Baselines are recomputed in the impact-window pipeline used for distillation; the cached-feature value for CLIP in Table~\ref{tab:lomo} is 0.227.\par}
\end{table}

Table~\ref{tab:truelomo} reports the outcome. Distillation improves LOMO by +0.104 over the single-frame baseline, with 15 of 17 materials improving (Wilcoxon $p = 0.002$), and by +0.053 over the four-frame baseline, with 14 of 17 improving ($p = 0.001$). The gains are uneven: the five largest (single-frame baseline $\to$ distilled) are on granular and natural surfaces, namely gravel $0.364 \to 0.708$, grass $0.377 \to 0.610$, dirt $0.285 \to 0.605$, leaf $0.290 \to 0.541$, and rock $0.247 \to 0.469$, which together account for about three-quarters of the total gain.

\subsection{Controls}
\label{sec:controls}

Table~\ref{tab:controls} collects the three controls, which together bound what this result can mean.

\begin{table}[t]
\centering
\small
\caption{\textbf{Controls, as pooled $\boldsymbol{\rho}$ on the held-out video split.} Correspondence arms hold the seed fixed, so the matched-teacher row is the first, and highest, of the three seeds (0.466) rather than their mean (0.440).}
\label{tab:controls}
\begin{tabular}{@{}llr@{}}
\toprule
\textbf{Control} & \textbf{Condition} & $\boldsymbol{\rho}$ \\
\midrule
\multicolumn{3}{@{}l}{\emph{How much is category recognition alone worth?}} \\
& CLIP material classifier (45.6\% acc.) $\to$ median $Q$ & 0.279 \\
& true material label $\to$ median $Q$ & 0.430 \\
\midrule
\multicolumn{3}{@{}l}{\emph{Is the trainable head doing the work? (cached single-frame features$^\dagger$)}} \\
& linear probe on $\log(Q+1)$ & 0.300 \\
& three-layer MLP on $\log(Q+1)$, no teacher & 0.275 \\
\midrule
\multicolumn{3}{@{}l}{\emph{Does the teacher need to correspond to the frame? (four frames + deltas)}} \\
& no distillation & 0.298 \\
& shuffled teacher (10.6\% material agreement) & 0.359 \\
& positional indexing (13.3\% material agreement) & 0.360 \\
& matched teacher & 0.466 \\
\bottomrule
\end{tabular}
\par\vspace{2pt}
{\footnotesize $^\dagger$The capacity pair is measured on the cached single-frame embeddings rather than the impact-window pipeline; the two rows are comparable with each other, and their values (0.275, 0.300) bracket the 0.281 single-frame figure.\par}
\end{table}

The first control asks how much of any $Q$ correlation is simply material recognition. Mapping the true material to its per-category median $Q$ gives $\rho = 0.430$ [0.400, 0.457], and a material classifier trained on CLIP features, at 45.6\% accuracy, mapped the same way gives $\rho = 0.279$ [0.246, 0.311]. Any pooled correlation near 0.28 is therefore indistinguishable from category recognition, which is precisely why we rest the argument on LOMO rather than on the pooled figure.

The second asks whether the trainable head is doing the work rather than the teacher. An identically shaped head given direct supervision on $\log(Q+1)$, with no audio involved, reaches only $\rho = 0.275$ [0.224, 0.293], below the 0.300 [0.267, 0.330] reached by a linear probe on the same features. Additional parameters buy nothing here: with roughly 14{,}000 noisy scalar targets, a regression head has little to learn from, whereas the teacher supplies a dense 1024-dimensional target whose geometry already encodes damping.

The third asks whether any structured target would do. Holding the seed fixed, distilling from correctly matched audio reaches $\rho = 0.466$ (the first, and highest, of the three seeds, whose mean is the 0.440 of Table~\ref{tab:distill}) against 0.298 with no distillation. Distilling from a fully shuffled teacher (10.6\% material agreement) still reaches 0.359, and positional indexing, used by an earlier version of our pipeline (13.3\% material agreement), reaches 0.360. Roughly a third of the raw gain therefore survives destroying the correspondence entirely, while the larger part, +0.107, requires the teacher to correspond to the frame it is paired with. Reported gains from cross-modal distillation should be read against a shuffled-teacher control for this reason.

\section{Discussion}

Three findings survive the controls. First, damping structure is present in audio and largely absent from single-frame vision, and separating the components of $Q$ shows that audio carries the damping term $\beta$ better than pitch, which is what licenses reading the transferred structure as physical. Second, a frozen visual encoder can acquire part of that structure from a frozen audio teacher without any $Q$ label entering training, and the gain holds on materials withheld from the projection head as well as the probe (0.234 to 0.338, 15 of 17 materials), where a category-level strategy earns nothing. Third, roughly half of the improvement needs no audio at all and follows from four frames instead of one, so part of what looks like a limit of vision is a limit of single-frame inputs.

The effect is modest in absolute terms. Most of the teacher's signal does not cross, and on the pooled measure the distilled encoder barely exceeds what perfect category knowledge would supply. That comparison is itself the methodological result we would most like carried forward: category identity alone reaches $\rho = 0.43$ on this task, so a pooled number near that value says little about physical understanding. Establishing physical transfer needs an evaluation under which category recognition earns nothing, namely true LOMO, together with a shuffled-teacher control (Section~\ref{sec:controls}).

The residual is a concrete target rather than a closed question. Damping depends on internal structure a camera often cannot observe, and the natural next step is $\beta$ itself, the component audio predicts best, nearly independent of $f_0$, and the one on which the distilled student already improves most.

\section{Limitations}
\label{sec:limitations}

Four caveats bound these results. First, evaluation is confined to Greatest Hits (977 videos, 17 materials) and to one visual domain, so whether the projection head learns a general appearance-to-damping mapping or associations tied to controlled drumstick footage is unresolved. Second, the $R^2 \geq 0.8$ filter keeps 63\% of annotated impacts at strongly material-dependent rates (Section~\ref{sec:extraction}), and the five largest per-material gains of Section~\ref{sec:truelomo} fall on materials retaining a third to just over a half of their impacts, so those correlations cover a selected minority of each category. The discarded impacts cannot be assessed with $Q$ at all, since there the target is undefined. Third, the four-frame model is evaluated on $Q$ alone (the $f_0$/$\beta$ probe in Appendix~\ref{app:decompose} uses a single-frame student), so although probing shows that audio carries $\beta$ better than $f_0$ (Section~\ref{sec:within}), we do not establish transfer of physical structure in general, and properties that impact sound encodes weakly or not at all, such as friction, may behave differently. Finally, we measure transferred structure by probing accuracy alone, so whether better $Q$ decodability yields better physical decisions remains to be shown on a task whose outcome is not a function of $Q$.

\bibliographystyle{plainnat}
\bibliography{references}

\begin{thebibliography}{25}
\providecommand{\natexlab}[1]{#1}
\providecommand{\url}[1]{\texttt{#1}}
\expandafter\ifx\csname urlstyle\endcsname\relax
  \providecommand{\doi}[1]{doi: #1}\else
  \providecommand{\doi}{doi: \begingroup \urlstyle{rm}\Url}\fi

\bibitem[Arandjelovi{\'c} and Zisserman(2017)]{arandjelovic2017look}
Relja Arandjelovi{\'c} and Andrew Zisserman.
\newblock Look, listen and learn.
\newblock In \emph{Proceedings of the IEEE International Conference on Computer
  Vision (ICCV)}, pages 609--617, 2017.

\bibitem[Aytar et~al.(2016)Aytar, Vondrick, and Torralba]{aytar2016soundnet}
Yusuf Aytar, Carl Vondrick, and Antonio Torralba.
\newblock Soundnet: Learning sound representations from unlabeled video.
\newblock In \emph{Advances in Neural Information Processing Systems
  (NeurIPS)}, volume~29, 2016.

\bibitem[Bear et~al.(2021)Bear, Wang, Mrowca, Binder, Tung, Pramod, Holdaway,
  Tao, Smith, Sun, et~al.]{bear2021physion}
Daniel~M Bear, Elias Wang, Damian Mrowca, Felix~J Binder, Hsiao-Yu~Fish Tung,
  RT~Pramod, Cameron Holdaway, Sirui Tao, Kevin Smith, Fan-Yun Sun, et~al.
\newblock {Physion}: Evaluating physical prediction from vision in humans and
  machines.
\newblock In \emph{Advances in Neural Information Processing Systems (NeurIPS)
  Datasets and Benchmarks Track}, 2021.

\bibitem[Chow et~al.(2025)Chow, Mao, Li, Seita, Guizilini, and
  Wang]{chow2025physbench}
Wei Chow, Jiageng Mao, Boyi Li, Daniel Seita, Vitor Guizilini, and Yue Wang.
\newblock {PhysBench}: Benchmarking and enhancing vision-language models for
  physical world understanding.
\newblock In \emph{International Conference on Learning Representations
  (ICLR)}, 2025.

\bibitem[Clarke et~al.(2023)Clarke, Gao, Wang, Rau, Xu, Wang, James, and
  Wu]{clarke2023realimpact}
Samuel Clarke, Ruohan Gao, Mason Wang, Mark Rau, Julia Xu, Jui-Hsien Wang,
  Doug~L James, and Jiajun Wu.
\newblock {RealImpact}: A dataset of impact sound fields for real objects.
\newblock In \emph{Proceedings of the IEEE/CVF Conference on Computer Vision
  and Pattern Recognition (CVPR)}, pages 1516--1525, 2023.

\bibitem[Gao et~al.(2023)Gao, Dou, Li, Agarwal, Bohg, Li, Fei-Fei, and
  Wu]{gao2023objectfolder}
Ruohan Gao, Yiming Dou, Hao Li, Tanmay Agarwal, Jeannette Bohg, Yunzhu Li,
  Li~Fei-Fei, and Jiajun Wu.
\newblock The {ObjectFolder} benchmark: Multisensory learning with neural and
  real objects.
\newblock In \emph{Proceedings of the IEEE/CVF Conference on Computer Vision
  and Pattern Recognition (CVPR)}, pages 17276--17286, 2023.

\bibitem[Gemmeke et~al.(2017)Gemmeke, Ellis, Freedman, Jansen, Lawrence, Moore,
  Plakal, and Ritter]{gemmeke2017audioset}
Jort~F. Gemmeke, Daniel P.~W. Ellis, Dylan Freedman, Aren Jansen, Wade
  Lawrence, R.~Channing Moore, Manoj Plakal, and Marvin Ritter.
\newblock {Audio Set}: An ontology and human-labeled dataset for audio events.
\newblock In \emph{IEEE International Conference on Acoustics, Speech and
  Signal Processing (ICASSP)}, 2017.

\bibitem[Girdhar et~al.(2023)Girdhar, El-Nouby, Liu, Singh, Alwala, Joulin, and
  Misra]{girdhar2023imagebind}
Rohit Girdhar, Alaaeldin El-Nouby, Zhuang Liu, Mannat Singh, Kalyan~Vasudev
  Alwala, Armand Joulin, and Ishan Misra.
\newblock {ImageBind}: One embedding space to bind them all.
\newblock In \emph{Proceedings of the IEEE/CVF Conference on Computer Vision
  and Pattern Recognition (CVPR)}, pages 15180--15190, 2023.

\bibitem[Gong et~al.(2023)Gong, Rouditchenko, Liu, Harwath, Karlinsky, Kuehne,
  and Glass]{gong2023cavmae}
Yuan Gong, Andrew Rouditchenko, Alexander~H Liu, David Harwath, Leonid
  Karlinsky, Hilde Kuehne, and James Glass.
\newblock Contrastive audio-visual masked autoencoder.
\newblock In \emph{Proceedings of the International Conference on Learning
  Representations (ICLR)}, 2023.

\bibitem[Gupta et~al.(2016)Gupta, Hoffman, and Malik]{gupta2016cross}
Saurabh Gupta, Judy Hoffman, and Jitendra Malik.
\newblock Cross modal distillation for supervision transfer.
\newblock In \emph{Proceedings of the IEEE Conference on Computer Vision and
  Pattern Recognition (CVPR)}, pages 2827--2836, 2016.

\bibitem[Hinton et~al.(2015)Hinton, Vinyals, and Dean]{hinton2015distilling}
Geoffrey Hinton, Oriol Vinyals, and Jeff Dean.
\newblock Distilling the knowledge in a neural network.
\newblock \emph{arXiv preprint arXiv:1503.02531}, 2015.

\bibitem[Huang et~al.(2023)Huang, Sharma, Xu, Ryali, Li, Li, Ghosh, Malik,
  Feichtenhofer, et~al.]{huang2023mavil}
Po-Yao Huang, Vasu Sharma, Hu~Xu, Chaitanya Ryali, Yanghao Li, Shang-Wen Li,
  Gargi Ghosh, Jitendra Malik, Christoph Feichtenhofer, et~al.
\newblock {MAViL}: Masked audio-video learners.
\newblock \emph{Advances in Neural Information Processing Systems},
  36:\penalty0 20371--20393, 2023.

\bibitem[Liu and Chen(2024)]{liu2024sonicsense}
Jiaxun Liu and Boyuan Chen.
\newblock {SonicSense}: Object perception from in-hand acoustic vibration.
\newblock In \emph{Conference on Robot Learning (CoRL)}, 2024.

\bibitem[Loshchilov and Hutter(2019)]{loshchilov2019adamw}
Ilya Loshchilov and Frank Hutter.
\newblock Decoupled weight decay regularization.
\newblock In \emph{International Conference on Learning Representations
  (ICLR)}, 2019.

\bibitem[Oquab et~al.(2024)Oquab, Darcet, Moutakanni, Vo, Szafraniec, Khalidov,
  Fernandez, Haziza, Massa, El-Nouby, et~al.]{oquab2023dinov2}
Maxime Oquab, Timoth{\'e}e Darcet, Th{\'e}o Moutakanni, Huy Vo, Marc
  Szafraniec, Vasil Khalidov, Pierre Fernandez, Daniel Haziza, Francisco Massa,
  Alaaeldin El-Nouby, et~al.
\newblock {DINOv2}: Learning robust visual features without supervision.
\newblock \emph{Transactions on Machine Learning Research}, 2024.

\bibitem[Owens and Efros(2018)]{owens2018audio}
Andrew Owens and Alexei~A Efros.
\newblock Audio-visual scene analysis with self-supervised multisensory
  features.
\newblock In \emph{Proceedings of the European Conference on Computer Vision
  (ECCV)}, pages 631--648, 2018.

\bibitem[Owens et~al.(2016{\natexlab{a}})Owens, Isola, McDermott, Torralba,
  Adelson, and Freeman]{owens2016visually}
Andrew Owens, Phillip Isola, Josh McDermott, Antonio Torralba, Edward~H
  Adelson, and William~T Freeman.
\newblock Visually indicated sounds.
\newblock In \emph{Proceedings of the IEEE Conference on Computer Vision and
  Pattern Recognition (CVPR)}, pages 2405--2413, 2016{\natexlab{a}}.

\bibitem[Owens et~al.(2016{\natexlab{b}})Owens, Wu, McDermott, Freeman, and
  Torralba]{owens2016ambient}
Andrew Owens, Jiajun Wu, Josh~H McDermott, William~T Freeman, and Antonio
  Torralba.
\newblock Ambient sound provides supervision for visual learning.
\newblock In \emph{Proceedings of the European Conference on Computer Vision
  (ECCV)}, pages 801--816, 2016{\natexlab{b}}.

\bibitem[Radford et~al.(2021)Radford, Kim, Hallacy, Ramesh, Goh, Agarwal,
  Sastry, Askell, Mishkin, Clark, Krueger, and Sutskever]{radford2021clip}
Alec Radford, Jong~Wook Kim, Chris Hallacy, Aditya Ramesh, Gabriel Goh,
  Sandhini Agarwal, Girish Sastry, Amanda Askell, Pamela Mishkin, Jack Clark,
  Gretchen Krueger, and Ilya Sutskever.
\newblock Learning transferable visual models from natural language
  supervision.
\newblock In \emph{International Conference on Machine Learning (ICML)}, 2021.

\bibitem[Riochet et~al.(2021)Riochet, Castro, Bernard, Lerer, Fergus, Izard,
  and Dupoux]{riochet2021intphys}
Ronan Riochet, Mario~Ynocente Castro, Mathieu Bernard, Adam Lerer, Rob Fergus,
  V{\'e}ronique Izard, and Emmanuel Dupoux.
\newblock {IntPhys} 2019: A benchmark for visual intuitive physics
  understanding.
\newblock \emph{IEEE Transactions on Pattern Analysis and Machine
  Intelligence}, 44\penalty0 (9):\penalty0 5016--5025, 2021.

\bibitem[Thankaraj and Pinto(2023)]{thankaraj2023sounds}
Abitha Thankaraj and Lerrel Pinto.
\newblock That sounds right: Auditory self-supervision for dynamic robot
  manipulation.
\newblock In \emph{Conference on Robot Learning}, pages 1036--1049. PMLR, 2023.

\bibitem[Tung et~al.(2023)Tung, Ding, Chen, Bear, Gan, Tenenbaum, Yamins, Fan,
  and Smith]{tung2023physionpp}
Hsiao-Yu Tung, Mingyu Ding, Zhenfang Chen, Daniel Bear, Chuang Gan, Josh
  Tenenbaum, Dan Yamins, Judith Fan, and Kevin Smith.
\newblock {Physion++}: Evaluating physical scene understanding that requires
  online inference of different physical properties.
\newblock \emph{Advances in Neural Information Processing Systems},
  36:\penalty0 67048--67068, 2023.

\bibitem[van~den Oord et~al.(2018)van~den Oord, Li, and Vinyals]{oord2018cpc}
Aaron van~den Oord, Yazhe Li, and Oriol Vinyals.
\newblock Representation learning with contrastive predictive coding.
\newblock \emph{arXiv preprint arXiv:1807.03748}, 2018.

\bibitem[Vyas et~al.(2025)Vyas, Chang, Yang, Huang, Gao, Richter, Chen, Le,
  Doll{\'a}r, Feichtenhofer, et~al.]{vyas2025pushing}
Apoorv Vyas, Heng-Jui Chang, Cheng-Fu Yang, Po-Yao Huang, Luya Gao, Julius
  Richter, Sanyuan Chen, Matt Le, Piotr Doll{\'a}r, Christoph Feichtenhofer,
  et~al.
\newblock Pushing the frontier of audiovisual perception with large-scale
  multimodal correspondence learning.
\newblock \emph{arXiv preprint arXiv:2512.19687}, 2025.

\bibitem[Yi et~al.(2020)Yi, Gan, Li, Kohli, Wu, Torralba, and
  Tenenbaum]{yi2019clevrer}
Kexin Yi, Chuang Gan, Yunzhu Li, Pushmeet Kohli, Jiajun Wu, Antonio Torralba,
  and Joshua~B Tenenbaum.
\newblock {CLEVRER}: Collision events for video representation and reasoning.
\newblock In \emph{International Conference on Learning Representations
  (ICLR)}, 2020.

\end{thebibliography}

\appendix

\section{Separate Probes for \texorpdfstring{$f_0$}{f0} and \texorpdfstring{$\beta$}{beta}}
\label{app:decompose}

Table~\ref{tab:decompose} probes the components of $Q$ separately, under the protocol of Section~\ref{sec:protocol}: ridge on frozen embeddings, Spearman $\rho$ on the held-out video split, and LOMO trained on 16 materials and evaluated within the held-out 17th. Targets are log-transformed as elsewhere. Across the 17,958 impacts, $Q$ is strongly rank-correlated with $f_0$ ($\rho = 0.87$) and moderately anti-correlated with $\beta$ ($\rho = -0.54$), while the two components are close to independent of each other ($\rho = -0.11$).

\begin{table}[h]
\centering
\small
\caption{\textbf{Probing $\boldsymbol{f_0}$ and $\boldsymbol{\beta}$ separately.} Audio embeddings predict the damping term $\beta$ better than the spectral term $f_0$, and the margin widens on unseen materials. Bold marks the better of $f_0$ and $\beta$ within each row and setting. The distilled row's LOMO columns use true LOMO, with the held-out material excluded from both the projection head and the probe, and comes from a single-frame student (the frame at impact), trained separately from the four-frame model of Tables~\ref{tab:distill}--\ref{tab:truelomo}, so it is compared like-for-like with the single-frame CLIP row.}
\label{tab:decompose}
\begin{tabular}{@{}l rr rr@{}}
\toprule
& \multicolumn{2}{c}{\textbf{Pooled $\boldsymbol{\rho}$}} & \multicolumn{2}{c}{\textbf{LOMO $\boldsymbol{\rho}$}} \\
\cmidrule(lr){2-3} \cmidrule(lr){4-5}
\textbf{Model} & $f_0$ & $\beta$ & $f_0$ & $\beta$ \\
\midrule
PE-AV (audio) & 0.700 & \textbf{0.729} & 0.485 & \textbf{0.686} \\
CAV-MAE (audio) & 0.667 & \textbf{0.732} & 0.423 & \textbf{0.719} \\
PE-AV (joint) & 0.671 & \textbf{0.682} & 0.439 & \textbf{0.628} \\
\midrule
CLIP & \textbf{0.347} & 0.221 & 0.137 & \textbf{0.257} \\
DINOv2 & \textbf{0.353} & 0.210 & 0.152 & \textbf{0.244} \\
PE-AV (video) & \textbf{0.396} & 0.337 & 0.134 & \textbf{0.207} \\
\midrule
CLIP distilled & \textbf{0.345} & 0.311 & 0.177 & \textbf{0.375} \\
\bottomrule
\end{tabular}
\end{table}

Two patterns are worth drawing out. First, for both audio encoders and the PE-AV joint embedding, $\beta$ is predicted better than $f_0$, and the margin grows under LOMO, which is where category identity stops carrying the prediction. Second, the visual encoders order the targets the other way when pooled, favoring $f_0$, which is consistent with pitch being partly inferable from visible object size, and that ordering reverses on unseen materials. Under true LOMO, the distilled student improves on both targets and improves most on $\beta$ ($0.257 \to 0.375$), so what transfers is preferentially the damping term rather than the spectral one.

\section{Within-Material \texorpdfstring{$Q$}{Q} Prediction}
\label{app:within}

Table~\ref{tab:within} gives the per-category breakdown summarized in Section~\ref{sec:within}. Correlations are computed within each material, so category identity carries no information.

\begin{table}[!ht]
\centering
\small
\caption{\textbf{Within-material $\boldsymbol{Q}$ prediction ($\boldsymbol{\rho}$),} shown for four representative rigid materials.}
\label{tab:within}
\begin{tabular}{@{}lrrrr@{}}
\toprule
\textbf{Model} & \textbf{Metal} & \textbf{Wood} & \textbf{Ceramic} & \textbf{Plastic} \\
\midrule
PE-AV (audio) & \textbf{0.67} & \textbf{0.57} & \textbf{0.31} & \textbf{0.45} \\
CAV-MAE (audio) & 0.65 & 0.45 & 0.12 & 0.32 \\
PE-AV (joint) & 0.64 & 0.41 & 0.02 & 0.34 \\
CAV-MAE (joint) & 0.43 & 0.25 & $-$0.10 & 0.18 \\
\midrule
PE-AV (video) & 0.36 & 0.07 & $-$0.08 & 0.00 \\
DINOv2 & 0.30 & 0.09 & 0.09 & $-$0.03 \\
CLIP$^\dagger$ & 0.06 & 0.03 & $-$0.14 & $-$0.19 \\
CAV-MAE (video) & $-$0.21 & $-$0.01 & 0.06 & $-$0.14 \\
\bottomrule
\end{tabular}
\par\vspace{2pt}
{\footnotesize $^\dagger$CLIP is measured on the impact-window pipeline with verified impact-to-embedding alignment, whereas the other rows come from the cached pipeline.\par}
\end{table}

\end{document}